\documentclass{article}

\usepackage[final]{neurips_2025_ml4ps}

\usepackage[utf8]{inputenc}
\usepackage[T1]{fontenc}
\usepackage{hyperref}
\usepackage{url}
\usepackage{booktabs}
\usepackage{amsfonts}
\usepackage{amsmath, amssymb}
\usepackage{nicefrac}
\usepackage{microtype}
\usepackage{xcolor}
\usepackage{graphicx}
\usepackage{amsthm}  % in the preamble
\usepackage{float} % in your preamble

\usepackage{subcaption}
\usepackage{makecell}
\title{Fast PINN Eigensolvers via Biconvex Reformulation}

\author{
  Akshay Sai Banderwaar \\
  School of Mechanical Sciences \\
  Indian Institute of Technology, Goa\\
  \texttt{banderwaar.sai.22063@iitgoa.ac.in}
  \And
  Abhishek Gupta \\
  School of Mechanical Sciences \\
  Indian Institute of Technology, Goa\\
  \texttt{abhishekgupta@iitgoa.ac.in}
}

\begin{document}

\maketitle

% === Abstract ===
\begin{abstract}
Eigenvalue problems have a distinctive forward-inverse structure and are fundamental to characterizing a system's thermal response, stability, and natural modes. Physics-Informed Neural Networks (PINNs) offer a mesh-free alternative for solving such problems but are often orders of magnitude slower than classical numerical schemes. In this paper, we introduce a reformulated PINN approach that casts the search for eigenpairs as a biconvex optimization problem, enabling fast and provably convergent alternating convex search (ACS) over eigenvalues and eigenfunctions using analytically optimal updates. Numerical experiments show that PINN-ACS attains high accuracy with convergence speeds up to 500$\times$ faster than gradient-based PINN training. We release our codes at \url{https://github.com/NeurIPS-ML4PS-2025/PINN_ACS_CODES}.
\end{abstract}

\section{Introduction}
Solving eigenvalue problems for differential operators involves finding non-trivial eigenvalue–eigenfunction pairs (eigenpairs) where the operator scales the function by its eigenvalue. Determining the function for a given eigenvalue resembles a forward problem, while extracting the eigenvalue from a function mimics an inverse problem. Physics-Informed Neural Networks (PINNs) have shown promise as high-resolution, mesh-free solvers for forward and inverse problems individually \cite{raissi2019physics, toscano2025pinns}, and these strengths could be brought to bear on the uniquely coupled forward–inverse structure of eigenvalue problems as well. However, existing PINN algorithms are known to be orders of magnitude slower than conventional numerical schemes \cite{mcgreivy2024weak, grossmann2024can}, primarily due to the difficulty of optimizing the highly non-convex PINN loss function. This loss combines terms representing the underlying differential equation and the initial and/or boundary conditions, which are often of incommensurable scale and can conflict under finite network capacity \cite{bischof2025multi}. Moreover, the initial flatness of the network's output---a symptom of the vanishing gradient problem \cite{wang2021understanding}---tends to draw PINNs towards trivial solutions \cite{wong2022learning, rohrhofer2023role}, an issue that is especially deleterious for eigenvalue problems.

Recent PINN methods for discovering eigenpairs rely on standard gradient-based iterative optimization \cite{ben2023deep, yoo2025physics, rowan2025solving, bonder2025pinns}, and therefore remain highly susceptible to conflicting loss terms and spurious basins of attraction in non-convex loss landscapes \cite{krishnapriyan2021characterizing, wong2501evolutionary}. One way to mitigate the conflict is to enforce boundary conditions directly into the network’s output, but applications have largely been limited to Dirichlet boundaries \cite{rowan2025solving, bonder2025pinns}. To reinforce the mesh-free flexibility of PINNs with accelerated convergence, this paper introduces a simple yet powerful trick: reformulating the PINN loss as a biconvex optimization problem. The key idea is to restrict training to the linear output layer of the PINN, keeping its hidden layers fixed to values initialized either randomly or according to some heuristic \cite{gallicchio2020deep}. For linear differential operators, this makes the physics residual linear in terms of the network's unknown parameters \cite{dong2023method, wang2024piratenets}. Consequently, the highly non-convex loss landscape is transformed into a biconvex one: convex (linear least-squares) in terms of the parameterized eigenfunction when the eigenvalue is fixed, and vice versa. We show that this structure admits a theoretically convergent alternating convex search (ACS) over eigenpairs, where each convex subproblem is solved optimally via analytical solutions. The overall framework not only yields substantial speedups compared to gradient-based optimization, but also improves accuracy through the analytic updates. The main contributions of this work are summarized below.
\begin{itemize}
    \item The differential eigenvalue problem is recast as a biconvex PINN optimization. Unlike prior PINN approaches, our reformulation accommodates varied boundary conditions and even applies to cases where the eigenvalue appears explicitly in the boundary operator (e.g., Euler’s buckling of a column with a free end).
    \item Alternating convex search for PINN optimization is proposed. Starting at a random initial eigenvalue estimate, ACS iteratively fixes one variable (the eigenvalue or the eigenfunction) and updates the other via analytic results. PINN-ACS converges rapidly and can be parallelized over a population of estimates for simultaneous discovery of multiple eigenpairs.
    \item We demonstrate the performance of our method on Euler's buckling problem, for solving Helmholtz equation on an L-shaped domain, and for obtaining the natural frequencies of a thin plate (governed by a fourth-order biharmonic equation). PINN-ACS achieves speedups exceeding 10$\times$, and up to 500$\times$, compared to gradient-based PINN training.
\end{itemize}

\section{Mathematical Preliminaries}
The general form of an eigenvalue problem for a differential operator defined over domain $\Omega$, with boundary $\partial\Omega$, can be written as:
\begin{subequations}\label{eq:eig_pb}
\begin{align}
\mathcal{D}(u_i)(\mathbf{x}) + \lambda_i h(u_i)(\mathbf{x}) &= 0, && \mathbf{x} \in \Omega, \label{eq:eig_pb_a}\\
\mathcal{B}(u_i)(\mathbf{x}) &= 0, && \mathbf{x} \in \partial\Omega. \label{eq:eig_pb_b}
\end{align}
\end{subequations}
% \begin{equation}
% \mathcal{N}(u_i)(\mathbf{x}) + \lambda_i h(u_i)(\mathbf{x}) = 0, \; \mathbf{x} \in \Omega,
% \end{equation}
% with boundary conditions specified along the edge $\partial\Omega$ of the domain:
% \begin{equation}
% \mathcal{B}(u_i)(\mathbf{x}) = 0, \; \mathbf{x} \in \partial\Omega.
% \end{equation}
\noindent Here, $\mathcal{D}(\cdot)$ is the differential operator, $h(\cdot)$ is a function, $\mathbf{x}$ represents the spatial coordinate, $\mathcal{B}(\cdot)$ is the boundary operator, $\lambda_i$ is the $i^{\mathrm{th}}$ eigenvalue and $u_i(\mathbf{x})$ is the corresponding eigenfunction.
In this work, we focus on linear self-adjoint operators, a common setting in recent studies and across many scientific applications \cite{ben2023deep,mishra2025eig}. Under this condition, the eigenfunctions are mutually orthogonal:
\begin{equation}
\langle u_i, u_j \rangle = \int_\Omega u_i(\mathbf{x})\, u_j(\mathbf{x})\, d\mathbf{x} = 0,  \; \text{for} \; i \neq j.
\end{equation}
% \begin{equation}
% \langle u_i, u_j \rangle = \int_\Omega u_i(\mathbf{x})\, u_j(\mathbf{x})\, d\mathbf{x} = 0,  \; \text{for} \; i \neq j.
% \end{equation}
%where $\langle \cdot, \cdot \rangle$ denotes the inner product over $\Omega$.
\section{Methodology}
\subsection{PINN Loss for Eigenvalue Problems}
Consider a multilayer perceptron parameterized by $\mathbf{\theta}$ to approximate the $i^{\mathrm{th}}$ eigenfunction as \( u_i(\mathbf{x}; \mathbf{\theta}) \). The PINN loss $\mathcal{L}(\lambda_i ,\mathbf{\theta})$ in terms of the model parameters and eigenvalue $\lambda_i$ can then be defined as:
\begin{align}
\operatorname*{arg\,min}_{\lambda_i, \, \mathbf{\theta}}\!
\; \mathcal{L}(\lambda_i ,\mathbf{\theta}) = 
&\int_{\Omega} \!\left| \mathcal{D} (u_i(\mathbf{x}; \mathbf{\theta})) 
    + \lambda_i h(u_i(\mathbf{x}; \mathbf{\theta})) \right|^2 d\Omega
+ \alpha_{\mathrm{BC}} \!\int_{\partial\Omega} \!\mathcal{B}(u_i(\mathbf{x}; \mathbf{\theta}))^2 d\left(\partial\Omega \notag \right) \\
&+ \alpha_{\mathrm{ref}} \!\left( u_i(\mathbf{x}^{\mathrm{ref}}; \mathbf{\theta}) - u_{\mathrm{ref}} \right)^2
+ \alpha_{\mathrm{ortho}} \!\sum_{j=1}^{i-1} 
    \left( \int_{\Omega} u_i(\mathbf{x}; \mathbf{\theta})\, u_j(\mathbf{x})\, d\Omega \right)^2.
\label{eq:total_loss}
\end{align}
The first term captures the error in satisfying the differential equation; the second term enforces the boundary conditions; the third term prevents a trivial solution by prescribing $u_{\mathrm{ref}} \neq 0$ at some reference point $\mathbf{x}^{\mathrm{ref}}$; the final term enforces orthogonality to eigenfunctions already discovered; $\alpha_{\mathrm{BC}}$, $\alpha_{\mathrm{ref}}$, $\alpha_{\mathrm{ortho}}$ control the trade-offs between components. The loss is usually evaluated numerically by computing the integrand at randomly sampled collocation points and summing the results. 

An alternate way to achieve solution non-triviality is to substitute the third term by a normalization condition: $\int_{\Omega} u_i(\mathbf{x}; \mathbf{\theta})^2\,d\Omega = 1$. However, this would induce nonlinear interactions between network parameters, hampering the biconvexification enabled by the form of Eq. (\ref{eq:total_loss}), discussed next.
%The above \textbf{loss function} comprises four terms: the first term represents the loss due to the partial differential equation (PDE), the second term enforces the boundary conditions, and the third term is a reference-point loss designed to prevent convergence to a trivial solution and fourth term represents orthogonality loss term to obtain multiple eigenpairs without converging to previously computed ones.
\subsection{Biconvex Reformulation }
In a depth-$L$ neural net, \(\theta = \{ W^{[l]}, \mathbf{b}^{[l]} \}_{l=1}^L\) and the transformation performed at the $l^{\mathrm{th}}$  layer is $a^{[l]}(\mathbf{x}) =\sigma^{[l]}(W^{[l]} a^{[l-1]}(\mathbf{x}) + \mathbf{b}^{[l]})$, where $a^{[0]}(\mathbf{x}) = \mathbf{x}$. While $\sigma^{[L]}$ is typically set to the identity function and $\mathbf{b}^{[L]} = \mathbf{0}$, the other non-linear activations, $\sigma^{[l]}$, render the network's output non-linear in terms of the hidden layers' weights and biases. This turns Eq. (\ref{eq:total_loss}) into a mathematical program with a highly non-convex loss landscape, posing stiff optimization challenges unique to PINNs \cite{wong2501evolutionary, sung2023neuroevolution}. 

\begin{figure}%[H] % H means "Here, exactly at this place"
    \centering
    \includegraphics[width=1.0\textwidth]{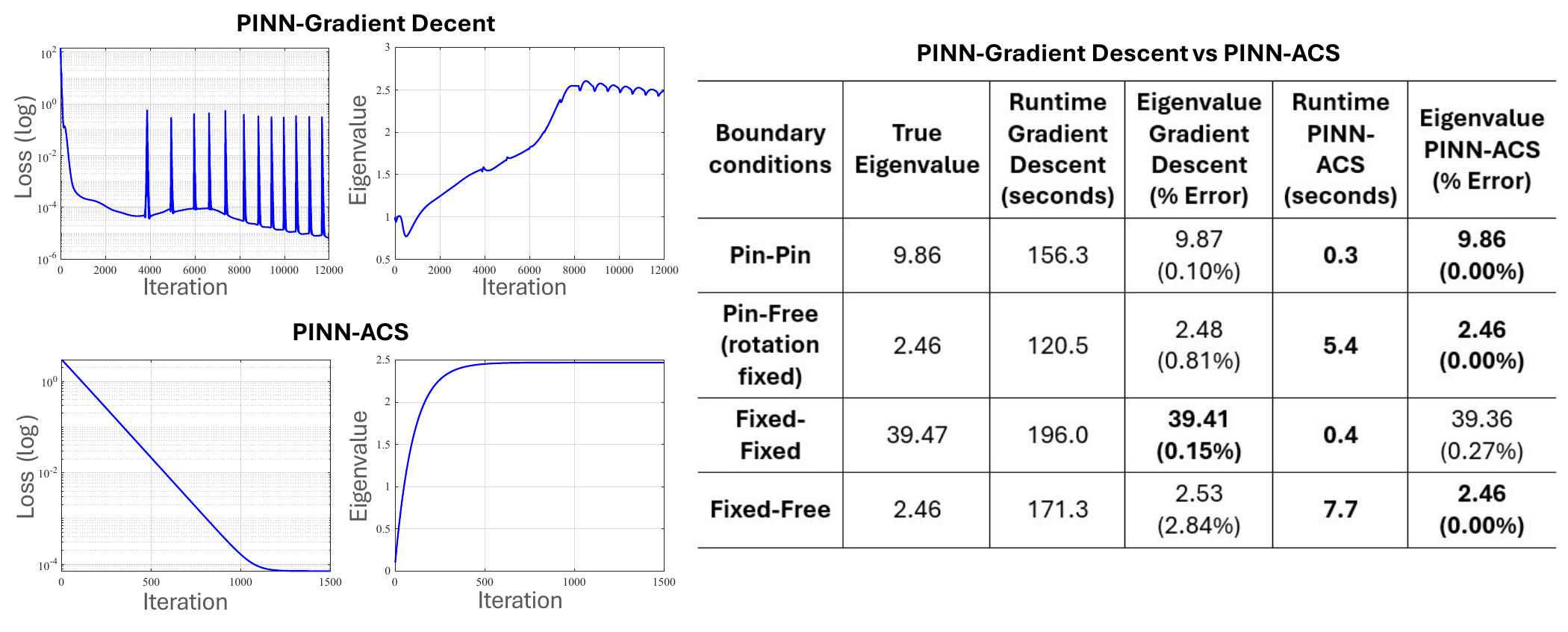}
\caption{Illustrative plots for Euler's buckling show that PINN-ACS leads to monotonic convergence as predicted by Theorem 1. Gradient descent using Adam \cite{yoo2025physics,kingma2014adam} oscillates due to conflicting loss terms. Tabulated average runtimes indicate substantial speedups with ACS alongside higher accuracy.}
\label{Fig1}
\end{figure}

In this paper, we reformulate the PINN eigensolver by restricting training to just the network's output layer. Hidden layers are either set to random values or according to some heuristic (e.g., by transferring weights from a related source problem) and are left untrained. Randomization preserves a network's universal approximation capacity under sufficient width and appropriate selection of random parameter distributions \cite{igelnik1995stochastic, li2023powerful}. As a result, we can write \( u_i(\mathbf{x}; \mathbf{\theta}) \) as \( u_i(\mathbf{x}; W^{[L]}) \) which is linear in terms of $W^{[L]}$. Plugging this simplified form of the eigenfunction into Eq. (\ref{eq:total_loss}) gives $\mathcal{L}(\lambda_i ,W^{[L]})$ where the third and fourth terms turn into linear least-squares expressions in $W^{[L]}$. Note that this would not be the case if the normalization condition was employed (as a substitute to the third term) for achieving non-trivial solutions. Although the first and (in certain cases) second terms give rise to non-linear interactions between $W^{[L]}$ and $\lambda_i$, conditioning them on a fixed $\lambda_i$ reduces the overall loss to a convex linear least-squares problem in terms of $W^{[L]}$ alone. Likewise, conditioning the loss on a fixed $W^{[L]}$ reduces it to a convex linear least-squares form in $\lambda_i$. The loss function is therefore "biconvexified" when jointly considering $\lambda_i, W^{[L]}$. Each convex subproblem admits a closed-form analytical solution by the Moore-Penrose pseudoinverse when the coefficients of $W^{[L]}$ or $\lambda_i$ are assembled into design matrices. In case of ill-conditioned design matrices, the closed-form solution obtained under Tikhonov regularization of $W^{[L]}$ provides a numerically stable approximation to the pseudoinverse (which is usually computed via singular value decomposition in most numerical libraries), and converges to it as the regularization coefficient approaches zero. 

\subsection{Alternating Convex Search}
Alternating convex search (ACS) is a robust algorithm for biconvex optimization \cite{gorski2007biconvex}. In the context of the eignevalue problem, each convex subproblem possesses an optimal analytical solution as discussed above. The alternating iterations—where one variable (either the eigenvalue or the parameterized eigenfunction) is fixed to its value from the previous step while the other is updated—are thus theoretically convergent and, in practice, converge rapidly.

\textbf{Theorem 1.} \textit{Let $\mathcal{L}^{(t)}$ be the loss function value for the differential eigenvalue problem after the $t^{\mathrm{th}}$ ACS iteration. Then the sequence $\{ \mathcal{L}^{(t)}\}_{t \in \mathbb{N}}$ generated by ACS converges monotonically.}
%$(\lambda,\theta) \in \Theta$, $\Theta \subseteq \mathbb{R} \times \mathbb{R}^n$ where n is the number of neurons, and let $\mathcal{L} : \Theta \to \mathbb{R}$ be bounded from below. For a fixed $\lambda_i$, the convex optimization problem $\min \{ \mathcal{L}(\lambda_i, \theta) \;|\; \theta \in \Theta_{\lambda_i} \}$, and for a fixed $\theta_{i+1}$, the convex optimization problem $\min \{ \mathcal{L}(\lambda, \theta_{i+1}) \;|\; \lambda \in \Theta_{\theta_{i+1}} \}$, are solvable. Then the sequence $\{ \mathcal{L}(\lambda_i,\theta_i) \}_{i \in \mathbb{N}}$ generated by ACS(Alternating convex search) converges monotonically.[\cite{gorski2007biconvex}]

\emph{Proof.} Squared terms in the loss function given by Eq. (\ref{eq:total_loss}) imply that its value must be greater than or equal to 0. Analytically optimal updates in each ACS iteration guarantee that $\mathcal{L}^{(t)}$ is monotonically decreasing. Since $\mathcal{L}$ is bounded from below, the sequence must converge to a limit value.

Empirical analysis shows that ACS typically converges to the eigenpair closest to its initial random eigenvalue estimate. By initializing a population of estimates, multiple ACS strands can be run in parallel to discover several eigenpairs simultaneously. The resulting eigenfunctions are incorporated into the fourth term of Eq. (\ref{eq:total_loss}) for the next generation of population-based search, thereby guiding each generation toward distinct eigenpairs. Given distributed hardware, this approach substantially reduces the time required to identify all eigenpairs within prescribed search bounds for the eigenvalues.

 \begin{figure}
    \centering
    \includegraphics[width=0.8\textwidth]{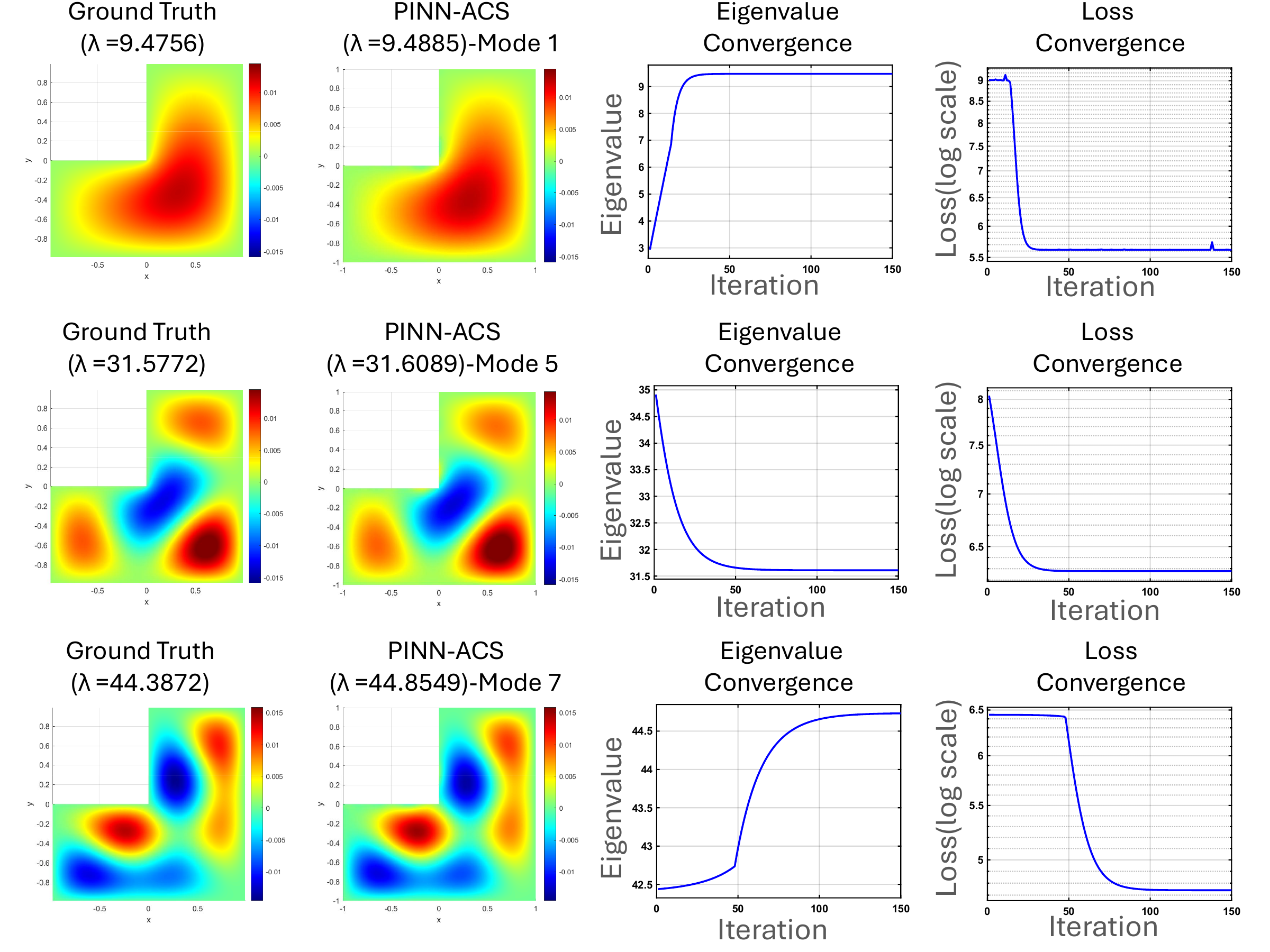}
    \caption{PINN-ACS closely approximates the ground truth (fine-mesh finite difference solutions) for the Laplace operator. Gradient descent results are omitted due to convergence difficulties within a reasonable time, especially for higher-order modes, consistent with observations in \cite{yoo2025physics}. In contrast, PINN-ACS achieves monotonic loss convergence in agreement with theory.}
    \label{Fig2}
\end{figure}

\section{Numerical Results}
 A shallow neural net with 500 neurons suffices for the 1D and 2D test problems considered in this paper. A shallow architecture allows all required derivatives of the network's output with respect to its inputs to be expressed analytically \cite{lagaris1998artificial}, thereby enabling efficient computation of higher-order derivatives without relying on automatic differentiation tools. Random Fourier features (cosine activations) were employed for effective learning of high frequency eigenfunctions \cite{tancik2020fourier}. Appropriate selection of the randomization bandwidth is important, with larger bandwidths facilitating the learning of high frequency patterns \cite{wong2022learning}. This selection could be automated by adapting the method from \cite{dong2022computing}. All codes for this work were developed in \textsc{MATLAB} R2023b taking advantage of its fast matrix computation capabilities, without GPU acceleration. The codes were run on a single 11th Gen Intel(R) Core(TM) i5-1135G7 processor with 8\,GB RAM $\sim$2.40\,GHz.

\subsection{Euler's Buckling}
The transverse deflection $w$ of a column under axial load is given by the fourth-order equation:
\begin{equation}
\frac{d^{4}w}{dx^{4}} + \lambda^{2}\frac{d^{2}w}{dx^{2}} = 0, 
\; x \in [0,1]. 
\end{equation}
The first eigenvalue of this problem produces critical buckling loads under varied boundary conditions, including unique cases where the eigenvalue appears in the boundary operator; e.g., zero shear condition at a free end is imposed by setting $\frac{d^{3}w}{dx^{3}} + \lambda^{2}\frac{dw}{dx} = 0$. Results comparing the performance of gradient-based PINN training using the formulation in \cite{yoo2025physics} and PINN-ACS are presented in Fig. (\ref{Fig1}). The key takeaway is that PINN-ACS is at least 10$\times$ faster than gradient descent. For the pin-pin and fixed-fixed cases, it achieves 500$\times$ speedup, while maintaining generally higher accuracy. The hidden model parameters were sampled uniformly at random from the interval $[-1, 1]$ for this example.

\subsection{Helmholtz Equation on L-shaped Domain}
The Helmholtz equation is the eigenvalue problem for the Laplace operator and is stated as:
\begin{equation}
\begin{aligned}
    \frac{\partial^2 w(x,y)}{\partial x^2} + \frac{\partial^2 w(x,y)}{\partial y^2} 
    &= -\lambda w(x,y), \; (x,y) \in \Omega.
\end{aligned}
\end{equation}
The problem is solved on an L-shaped domain with Dirichlet boundaries, i.e., $w(x,y) = 0, (x,y) \in \partial \Omega$. Comparisons to a classical numerical scheme are shown in Fig. (\ref{Fig2}), with PINN-ACS achieving a high degree of accuracy for both low- and high-frequency eigenfunctions. The hidden parameters of the model were sampled uniformly at random from the interval $[-8\pi, 8\pi]$ for this example.

% remove the global \color{blue} from preamble!
 \begin{figure}
    \centering
    \includegraphics[width=0.85\textwidth]{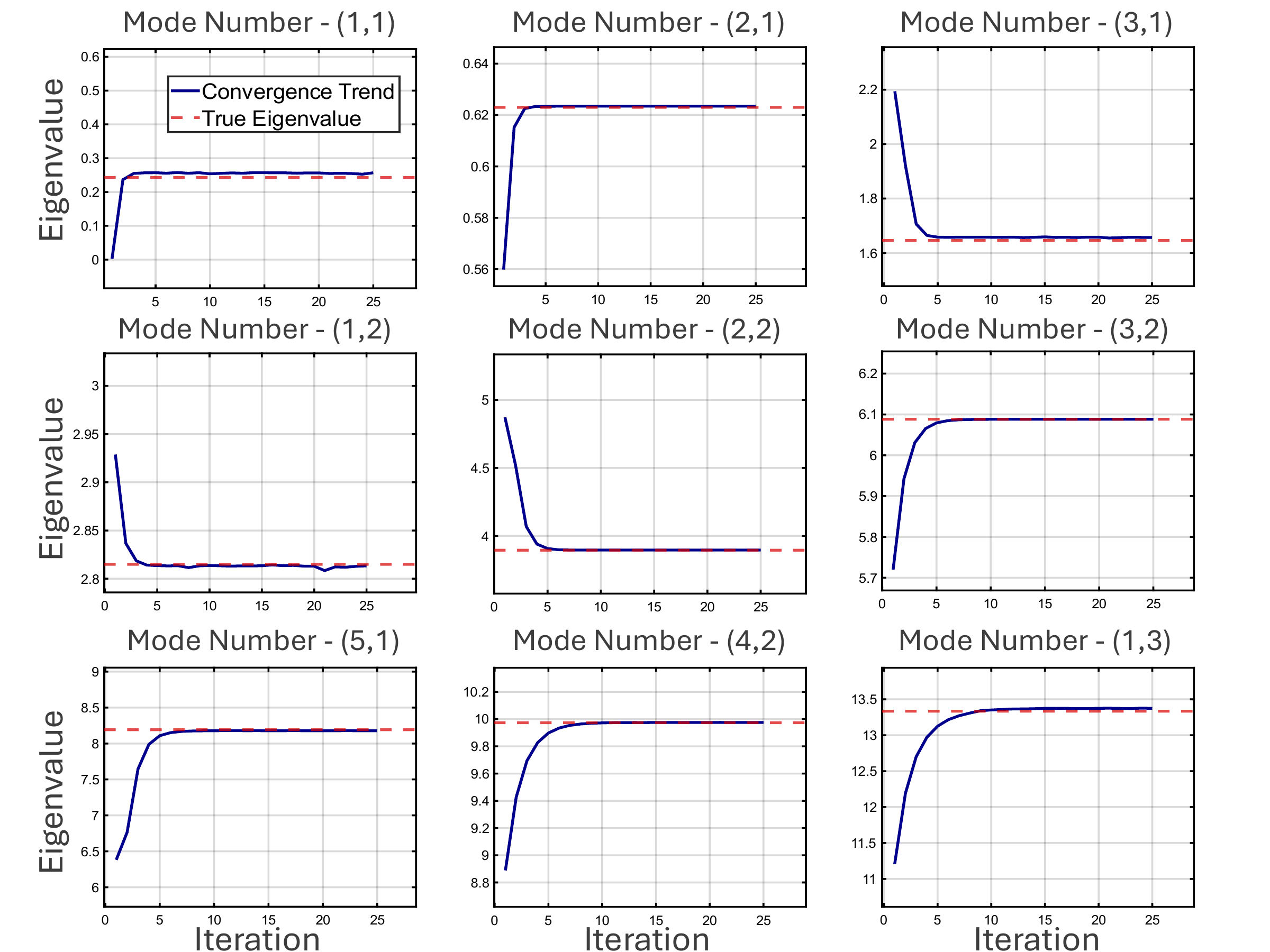}
    \caption{Convergence trends of PINN-ACS eigenvalues to the ground truth for free vibrations of a thin plate \cite{timoshenko1959theory}. Gradient descent results excluded as it fails to converge reliably for higher modes.}
    \label{Fig3}
\end{figure}

\subsection{Biharmonic Equation for Thin Plates}
The vibration of thin plates is governed by the following biharmonic eigenvalue problem:
\begin{equation}
\frac{\partial^4 w}{\partial x^4}
+ 2 \frac{\partial^4 w}{\partial x^2 \partial y^2}
+ \frac{\partial^4 w}{\partial y^4}
= \lambda^4 w, \quad (x, y) \in [0,10]\times[0,5]. \label{eq:kirchhoff}
\end{equation}
    The problem is solved for a rectangular plate under simply supported boundary conditions, assuming Poisson's ratio $\nu=0.3$. The PINN's hidden parameters were sampled uniformly from the interval $[-\pi,\pi]$. Fig. (\ref{Fig3}) depicts the smooth convergence of the estimated eigenvalues to the ground truth.
%\begin{align}
%& w(x_1, x_2, t) = 0 \quad \text{at} \quad x_1 = 0,a \quad \text{and} \quad x_2 = 0,b \\[1.5ex]
%& \left( \frac{\partial^2 w}{\partial x_1^2} + \nu \frac{\partial^2 w}{\partial x_2^2} \right) = 0 \quad \text{at} \quad x_1 = 0,a \\[1.5ex]
%& \left( \frac{\partial^2 w}{\partial x_2^2} + \nu \frac{\partial^2 w}{\partial x_1^2} \right) = 0 \quad \text{at} \quad x_2 = 0,b
%\end{align}

\section{Conclusions}
This paper fills a gap in PINN applications by introducing a fast PINN eigensolver based on a biconvex loss reformulation. The alternating convex search (ACS) procedure is theoretically guaranteed to achieve monotonic convergence for biconvex problems, and provides substantial speedups in practice compared to gradient-based PINN training. Future work will extend the approach to nonlinear operators and larger-scale physics applications. One way to address nonlinearity is by the Picard method \cite{paniconi1994comparison} where the problem is iteratively linearized at the differential equation level using previous solution approximations, thereby preserving applicability of the PINN-ACS algorithm on alternating linear least-squares subproblems. More broadly, ACS provides a general framework for solving inverse problems where both unknown field functions and parameters are simultaneously sought. 

%We proposed a biconvex optimization framework for solving eigenvalue problems using Modified Physics-Informed Neural Networks (PINNs), addressing key limitations of standard gradient-based training. we validated on two benchmark problems. First, in the buckling problem, we tested five different boundary conditions and demonstrated that our approach consistently achieved significantly faster convergence compared to gradient-descent-based PINNs. Second, in the case of the Laplacian operator on an L-shaped domain, we successfully discovered multiple eigenpairs. For this problem, we did not include a direct comparison with gradient descent, since it was already shown how slowly it converges to the first eigenmode—which is often crucial in buckling analysis—and thus is not worthwhile choice for problems requiring multiple eigenmodes.

%For higher modes, where nodal-line sensitivity makes reference-point selection critical, we addressed this by sampling multiple candidates and retaining the one with minimum loss. Future work may focus on eliminating reference-point dependence entirely by enforcing normalization constraints (e.g., a unit $L^2$ norm), while maintaining biconvexity to further accelerate convergence. Finally, although the present formulation assumes linear differential operators, extending the framework to nonlinear operators represents a promising direction.  
\section*{Acknowledgements}
This work was supported by the Ramanujan Fellowship from the Anusandhan National Research Foundation, Government of India (Grant No. RJF/2022/000115).

% === References ===
\bibliographystyle{unsrt}
\bibliography{references}

\end{document}